\documentclass[conference]{IEEEtran}
\usepackage{graphicx}
\usepackage{epstopdf}
\begin{document}
%

\title{Offshore Wind Farm Layout Optimization Using Adapted Genetic Algorithm: A different perspective}

\author{\IEEEauthorblockN{Feng Liu, Zhifang Wang}
\IEEEauthorblockA{Electrical and Computer Engineering\\
Virginia Commonwealth University,Richmond,VA 23220\\
Email: liuf2@vcu.edu, zfwang@vcu.edu}
}

\maketitle

\begin{abstract}
test In this paper we study the problem of optimal layout of an offshore wind farm to minimize the wake effect impacts. Considering the specific requirements of concerned offshore wind farm, we propose an adaptive genetic algorithm (AGA) which introduces location swaps to replace random crossovers in conventional GAs. That way the total number of turbines in the resulting layout will be effectively kept to the initially specified value. We experiment the proposed AGA method on three cases with free wind speed of 12 m/s, 20 m/s, and a typical offshore wind distribution setting respectively.  Numerical results verify the effectiveness of our proposed algorithm which achieves a much faster convergence compared to conventional GA algorithms.
\end{abstract}


\IEEEpeerreviewmaketitle

\section{Introduction}
As the costing of wind generation declines, combined with the uncertainties of fossil fuel price and the incentives by the governmental efforts to reduce the greenhouse effects and to address other environmental concerns, more and more wind farms all over the world are being constructed or to be granted by their respective governments. Wind energy is an established source of renewable energy and can complement fossil fuel derived energy in some way, especially in this era when manufacture technologies make wind turbines function more reliably and endurably.

Important data showcase that the installed wind energy capacity reached 196,630 MW on worldwide while claiming an increase of 37,642MW in the single year of 2010 \cite{citation1}. Of all the interesting topics regarding the wind energy investments like increasing reliability and efficiency of wind turbines, optimizing the maintenance, assembly and installation of offshore and onshore turbines and their substructures, and improving the design and layout of wind farms, the wind farm layout design is one of the most important because the most profitable design has to be found out in order to gain optimal output of the wind farm given certain amount of investment. An unsophisticated design will decrease the wind power capture and increase the maintenance cost. Because of the presence of the wake effect, the total output of the wind farm is not just a simple summation of each wind turbine's rated power. In fact,it is often less than the sum of each wind turbine. As a result, wind farm layout design could be defined as a Wind Farm Layout Optimization Problem (WFLOP) which aims at achieving higher aggregate efficiency of converting wind energy to electricity.

 According to Claus Madsen from ABB company, ``80\% of the population (in United states) live along the shore line, 80\% of the energy from unshore wind is in the middle of the country, get the energy from the middle of the country to the shore lines is a big challenge very expensive. That's why offshore wind located near the shore line is the right technology. This fact is also true for China, who became number one in total installed capacity and accounts for more than 50 \% of the world market for new wind turbines in 2010, and the rest of the world. By 2010, 80 per cent of people will live within 60 miles of the coast \cite{citation2}. Further more, offshore wind conditions are favorable compared to wind farms on land, with stronger and steadier wind speeds. All the aforementioned facts spur the research as well of the development of offshore wind farms. In this paper, we will not consider the electrical cable layout, the installation and maintenance of the offshore wind farm, the optimal layout of the wind turbines is the main focus of this paper.

As early as 1994, Mosetti et al first presented the wind farm layout optimization problem using genetic algorithm \cite{RefWorks:8}. Later Grady et al studied the same problem with more individuals in a generation and evolve more 3000 generations to get the optimal result \cite{RefWorks:12}. In recent years, Eroglu and Seckiner discussed about the WFLOP using the particle filtering approach \cite{RefWorks:112}, in which three different cases are optimized using the particle filtering approach, and the result can compete with ant colony and evolutionary strategy algorithms.  Emani and Pirooz studied the same problem using genetic algorithms \cite{RefWorks:9}, and some improvement is achieved by taking into consideration of comparative weights of the cost of the whole wind farm each year and the yearly energy output. Bilbao and Elba \cite{RefWorks:10} used the simulating annealing method to achieve the maximum of the wind farm output. Perez et al use mathematical programming techniques to study the offshore wind farm layout problem \cite{P2013389}, and test their proposed procedure using German offshore wind farm Alpha Ventus.

However, most, if not all, of the genetic algorithms in the references use crossover operation, which is a way to heuristically search the optimal solution, rather than trying to make the the individual's fitness value increase on purpose. What's more the crossover operation doesn't have a physical meaning behind it in our wind farm optimization problem. What's more, the traditional evolution algorithm usually converges to a local optimum. Although the inclusion of chaos in the genetic algorithm cannot fully solve the problem, it can improve the computation efficiency and avoid the local minimum by introducing more randomness in the algorithm \cite{citation3}.
By extracting energy from the wind, a turbine creates a cone (wake) of slower and more turbulent air behind it. The wake model has been studied by several authors in the fluid-aerodynamic field. In this paper, we use Jensen wake model to compute the wind speed reduction in the far wake case.
In this paper, we propose a new method called Adaptive Genetric Algorithm (AGA) to solve the concerend offshore  WFLOP which has special requirements of total rated output power. Three cases are experimented to demonstrate the performance of our algorithm.

\section{Problem formulation}
In this section, we first present the wake model, then give the detailed implementation of the proposed adaptive genetic algorithm for WFLOP. When the wind direction is changed $\theta$ degree from the reference direction clockwise, the cartesian axis should also rotate $\theta$ degree and the new coordinates can be calculated using
\begin{equation}
L(\theta ) = \left[ \begin{array}{l}
x(\theta )\\
y(\theta )
\end{array} \right] = \left[ {\begin{array}{*{20}{c}}
{\cos \theta }&{ - \sin \theta }\\
{\sin \theta }&{\cos \theta }
\end{array}} \right]L(0)
\end{equation}
where $L(0)$ is the original coordinates of the a wind turbine. For notation simplicity, we use $[x,y]$ to represent $[x(\theta),y(\theta)]$ when no confusion occurs.

\subsection{Wake model}
As a wind turbine acts to withdraw momentum from the wind, it captures part of the kinetic energy in the wind stream and leaves behind it a wake with the wind speed reduced. The electric energy output of another turbine operating inside this wake will therefore be affected.  The assumption of wake model in this paper is the same with Mosetti et al \cite{RefWorks:8}. Here, a simplified wake decay model developed by N. O. Jensen \cite{RefWorks:14} is employed, as it offers a best balance between positive and negative prediction errors.

For a turbine $i$, the wind speed before turbine $i$ can be calculated by
\begin{equation}
{u_i} = v(1 - {D_{vi}}),
\end{equation}
where $v$ is the free wind speed and ${D_{vi}}$ the velocity deficiency of the turbine $i$.
\begin{equation}
{D_{vi}} = \sqrt {\sum\limits_{j \in {\Omega_i}} {{D_{vij}}^2}},
\end{equation}
where ${D_{vij}}$ is the velocity deficiency of the turbine $i$ incurred by the wake of the turbine $j$. ${\Omega_i}$ is the set that contains all the turbines which have wake impact on the turbine $i$ and can be defined as ${\Omega _i} = \{  j| d_{ij}={y_j} -{y_i}>0 \textrm{ and } {A_{ij}} > 0\}$. $D_{vij}$ is defined by
\begin{equation}
{D_{vij}}{\rm{ = }}\frac{{1{\rm{ + }}\sqrt {1 - {C_{Tj}}} }}{{{{(1 + {\raise0.7ex\hbox{${k{d_{ij}}}$} \!\mathord{\left/
 {\vphantom {{k{|d_{ij}| }} {{R_i}}}}\right.\kern-\nulldelimiterspace}
\!\lower0.7ex\hbox{${{R_i}}$}})}^2}}} \cdot \frac{{{A_{ij}}}}{{{A_i}}},
\end{equation}
where ${C_{Tj}}$ is the thrust coefficient of the turbine $j$, ${A_i}$ is the rotor area of the turbine $i$, ${A_{ij}}$ is the intersection area between the wake of turbine $j$ and ${A_{i}}$, $k$ is the decay factor that can be calculated using $k = 0.5/\ln (h/{z_0})$. Here, $h$ is the turbine height, $z_0$ is the surface roughness coefficient.

In order to calculate the intersection area ${A_{ij}}$, we define $r_j$ as the radius of the wake incurred by turbine $j$ just before the turbine $i$, $R_i$ is the rotor radius of turbine $i$ and ${x_{ij}}$ as the horizontal distance between turbine $i$ and $j$, which is the projection of the distance of $i$ and $j$ on a line that is perpendicular to the wind direction. The intersection area ${A_{ij}}$ is calculated by the following formula,
\begin{equation}
A_{ij}=\left\{ {\begin{array}{*{20}{ll}}
{ 0, }&{  x_{ij}> {r_j} + {R_i} }\\
{\pi R_{_i}^2, } &{ x_{ij}<{r_j} - {R_i}  }\\
{ A_s, } & {{r_j} < {x_{ij}} < {r_j} + {R_i}}\\
{ A_b, } &{ r_j - R_i < x_{ij} < r_j }
\end{array}} \right.
\end{equation}

When ${x_{ij}}$ satisfies ${r_j} < {x_{ij}} < {r_j} + {R_i}$, it means the center of the rotor circle is outside of the wake of turbine $j$ with a wake radius to be $r_j$. The intersection area ${A_s}$ can be calculated using the following formula. For notation simplicity, denote $a = {r_j}$, $b = {R_i}$, $c = {x_{ij}}$
Let $\alpha  = \arccos \frac{{{a^2} + {c^2} - {b^2}}}{{2ac}}$,
${A_1} = \frac{\alpha }{{{{180}^0}}}\pi {a^2}$, $\beta  = \arccos \frac{{{c^2} + {b^2} - {a^2}}}{{2bc}}$, ${A_2} = \frac{\beta }{{{{180}^0}}}\pi {b^2}$, ${A_3} = ac\sin \alpha $.
Then ${A_s} = {A_1} + {A_2} - {A_3}$.
Similarly, when ${r_j} - {R_i} < {x_{ij}} < {r_j}$ is satisfied, the intersection area  ${A_b}$ can be calculated similarly. Let $\alpha  = \arccos \frac{{{a^2} + {c^2} - {b^2}}}{{2ac}}$, and ${A_1} = \frac{\alpha }{{{{180}^0}}}\pi {a^2}$, $\beta  = {180^o} - \arccos \frac{{{c^2} + {b^2} - {a^2}}}{{2bc}}$, ${A_2} = \frac{\beta }{{{{180}^0}}}\pi {b^2}$, ${A_3} = ac\sin \alpha$, ${A_b} = \pi {c^2}-({A_1}-{A_2}-{A_3})$

\subsection{Optimization of the wind farm layout}\label{sec.wind_opt}
\begin{figure}
\begin{center}
\includegraphics[width=3.1in,height = 2in]{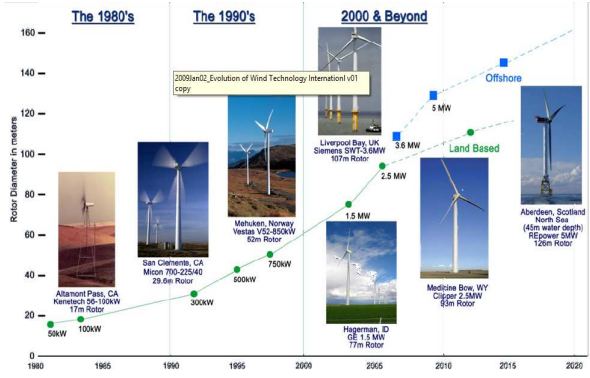}
\caption{Evolution of commercial wind technology (NREL)}
\label{trend}
\end{center}
\end{figure}

This section discusses how to determine the total number of wind turbines in a wind farm given specific condtions as follow. The concerned layout problem of wind turbines in the offshore wind farm is based on the settings of the recent project in Virginia Beach granted to Dominion Virginia Power in October 2013.  With a nameplate capacity of 2,000 megawatts (MW), it will be the world's largest offshore wind farm. All the wind turbines will be erected on nearly 113,00 acre offshore area, which is about a 20km by 20 km square. As can be seen from statistics \cite{farmlist}, there exists a trend to favor high-power wind turbines  in large offshore wind farms, as is illustrated in Fig.\ref{trend} given by NREL.

%\subsubsection{unique type or turbines or mixed types of wind turbines}
In this paper, we assume that only one type of wind turbines will be be erected over the wind farm. When more than one types of turbines are adopted with different hub heights, rotor diameters, and rated power etc, the construction and maintenance costs could be highly increased. Because the workers and operators have to be trained elaborately in order to construct and operate each specific type of wind turbines at different designated locations correctly. What's more, the maintenance will be more expensive due to the higher cost of repair and replacement of variety parts and components.  Therefore we assume that the same type of wind turbines with the rated power of $p^{\max}=5$MW  will be employed.

\subsubsection{fixed or flexible number of wind turbines}
Since Mosetti's paper \cite{RefWorks:8} presents the popular aggregate cost function of $C_{tot}(N)=N(2/3+1/3e^{-0.00174N^2})$, most related literatures set the total number of the wind turbines, $N$, as one of the optimization variables, with the optimization objective function of $\max {P_{total}(N)}/C_{tot}(N)$ where $P_{total}$ is the aggregate turnbine output power as defined in eq.(6). However, when it comes to the planning problem of a large wind farm with the targeted $N$ greater than 50, the total cost will approximate a linear function , i.e. $C_{tot}(N) = 2/3 N$. And the corresponding objective function now changes to  $\max { P_{total}(N)}/N$, that is, equivalent to maximize the average unit output power, which in fact prefers smaller $N$ due to the fact that the larger $N$ the more wake effect impact and the smaller average unit tubine's output power. Considering the rated total output power constraints of the wind farm which requires that $N\ge N_0$ (with $N_0>50$), the optimization setting will drive the best setting of turnbine number to be $N = N_0$. Hence the original optimization can be formulated with a fixed number of turbines and its optimizaiton function now turns into simply maximize the total output power $\max P_{total}(N_0)$.

However, according to the authors' best knowledge, the assumption of maximum of $1/3$ reduction of unit cost presented by Mosetti is not validated by real world data or experiment results. %{\bf my understanding is this is pretty  reasonable approximation, with the unit price drops to a constant value as total number increases. }
%{\bf the following part is not very clear}

\begin{figure}
\begin{center}
\includegraphics[width=2.5in]{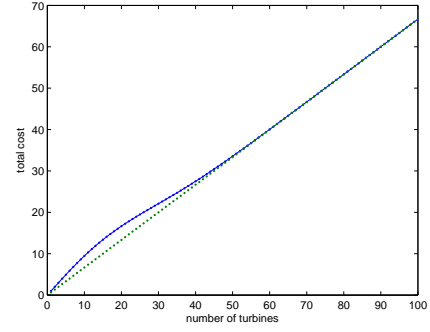}
\caption{cost function vs y=2/3N}
\label{costplot}
\end{center}
\end{figure}

Even if it holds true, it is not useful for a large wind farm because the large offshore wind farm are comprised of hundreds of the wind turbines. As can be seen from Fig.\ref{costplot}, the cost function curve approaches the linearly increasing cost function of $2/3N$ when $N>50$. That means the inclusion of the cost in the objective function will not affect so much the final optimal solution when $N>50$ compared to when $N$ is chosen from rather small numbers $(<20)$.

Therefore in this paper we assume that the total number of wind turbines $N$ should be a fixed value rather than a variable to be optimized. The number $N$ can be calculated using the nameplate output of the farm divided by rated power of each wind turbine.
As in the example in Virginia Beach, the number of the wind turbines should be $2000/5=400$. For the convenience of demonstration, we shrink the problem of placement of 400 turbines over a 20km by 20km area to a smaller size site with the corresponding number of wind turbines. Here, we choose to place 16 turbines on a 4km by 4km wind farm which has the same average density of wind turbines with the original problem.

\subsubsection{reduction of solution space using the fixed number of turbines}
\label{sectref}
With the fixed number of turbines to be optimized over an area, we find the computation complexity will be greatly reduced. Divide the 4km by 4km area into 20 by 20 cells, with each cell width to be 200m. Like Ref.\cite{RefWorks:15}, all the intersection points are the possible locations. There are $441$ possible locations.
Unlike Ref.\cite{RefWorks:15}, our wind farm size is assumed based on the practical derivation. However, in Ref.\cite{RefWorks:15}, the results show that about 15 to 20 turbines are placed in an area of 500m by 500m, which is a different case in offshore wind farm. The diameter of offshore wind turbines are usually larger than 100m, making the optimized area too crowded taking into consideration of the intensive turbulence impact on the wind turbines. If the number of wind turbine is unset, the solution space will be $2^{441}$ possible solutions, which is approximately $5.68 \times {10^{132}}$. However, if the $N$ is set to be 16, which is an n choose k problem, the solution space is $3.49 \times 10^{28}$. The solution space is greatly reduced.

\subsubsection{optimization objective function:}
\begin{figure}
\begin{center}
\includegraphics[width=2.5in]{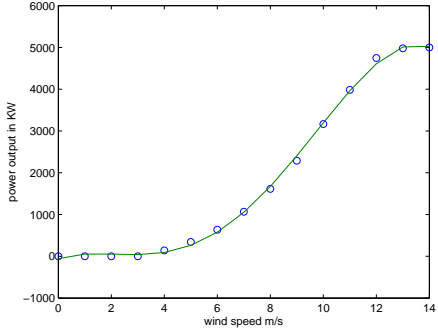}
\caption{Wind turbine output power $p_g(v)$ with $p^{\max} = 5$MW}
\label{powerfit}
\end{center}
\end{figure}

The optimization goal function is to maximize the total power output of the wind farm, and can be described by
\begin{equation}\label{eq.P_{total}}
{P_{total}} = \sum\limits_v {\sum\limits_\theta  {{f_w}(\theta ,v)} {P_{G|\theta ,v}}},
\end{equation}
where ${f_w}(\theta ,v)$ is the two dimensional possibility distribution function of wind over wind speed $v$ and wind direction $\theta$. In this paper, we divide $360^o$ into 12 sections, with each representing $30^o$. $P_{G|\theta ,v}$ is the total power generated under the condition of wind speed to be $v$ and direction to be $\theta$,
\begin{equation}
{P_{G|\theta ,v}} = \sum\limits_{i = 1}^N {{P_{Gi|\theta ,v}}} ,
\end{equation}
where $N$ is the number of turbines. For notation simplicity, we will use $P_G$ instead of $P_{G|\theta ,v}$ when no confusion occurs.
The power generated by turbine $i$ is ${P_{Gi}} = {p_g}({u_i})$, $u_i$ is the wind speed before turbine $i$ after taking into account the wake effect impacts. ${p_g}(\cdot)$ is the power output which is given by the wind turbine manufacture. Here we use the power curve given by \cite{powercurve}. When the wind speed is larger than $14m/s$ and smaller than the cut out speed, the power output remains constant as 5MW. A polynomial approximation of $p(v)= -0.9114v^4+21.6654v^3 -113.1189v^2+201.1211v -55.0267$ is used to fit the discrete data provided by the manufacture when the wind speed is smaller than the cut-off speed 14$m/s$.
Hence the power curve can be described by the following function and Fig.\ref{powerfit} shows a typical output power curve.

\begin{equation}
{p_{g}}(v) = \left\{ {\begin{array}{*{20}{ll}}
{0, } & {v < 3{\mkern 1mu} }\\
{p(v),} & {3 \le v < 14{\mkern 1mu} }\\
{p^{\max},}  & {v \ge 14{\mkern 1mu} }
\end{array}.} \right.
\end{equation}
Therefore the efficiency of the wind farm layout can then be derived as follows:
\begin{equation}
\eta  = \frac{{{P_{total}}}}{N{\sum\limits_v {\sum\limits_\theta  {{f_w}(\theta ,v){p_g}(v)} } }}.
\end{equation}

\section{Adaptive Genetic Algorithm}
Conventional genetic algorithms can't be adopted directly because its initialization, crossover and mutation procedures cannot guaranttee the desired total number of turbines. In our problem, each individual contains 441 bits with 0 representing no turbine at the location and 1 means there is a wind turbine at that point. Note that only 16 bits whose value are 1s are spread out in the 441 bits, making the initialization different from the general randomly initialize the individual. Also, crossover is not validated because its result may lead to the change of number of turbines in each individual. For example, cross over $P_1$ ${\rm{[1 0 1 0 0 1 0 1 0 1]}}$ and $P_2$ ${\rm{[1 1 1 1 0 1 0 0 0 0]}}$ from the 6th bit results in ${\rm{[1 0 1 0 0 1 0 0 0 0]}}$ and ${\rm{[1 1 1 1 0 1 0 1 0 1]}}$. The former one has 3 turbines over the ten possible points, while the latter has 7 turbines, which is conflict with our fixed number of turbines. The mutation procedure is also adaptive to make the 1s in each individual unchanged, it is done by mutate twice by replacing 1 into 0 or vice versa. Last but not least, we incorporate another operation that can boost the heuristic searching procedure: the relocation of the least efficient turbine.
The detailed AGA can be describe by the following procedure:

Step 1: Initialization. Using the chaotic map $x_{n+1} = 4x_{n}(1 - x_{n})$ to generate values between zero and one. Multiply $x_{n+1}$ by $N$ and round it to the nearest integer, the integer is the location of the first turbine. Continue the procedure until all $N$ turbines are generated randomly. Generate $P_N$ individuals.

Step 2: Evaluation and reordering: Evaluate the fitness value of each individual and order the individuals according to their fitness value in descending order.

Step 3: Elites selection: The first $P_I$ individuals are the elite ones, and are copied to the next generation directly.

Step 4: Relocation of the worst turbine: Use the first $P_I$ individuals to generate $P_D$ descendants by relocating the least efficient wind turbines randomly using chaotic map.

Step 5: Inclusion of aliens: Include the $P_A$ aliens as the descendants by generating randomly using chaotic map like the initialization step.

Step 6: Mutation: randomly choose 1 bit out of $M$ bits from the best individual and toggle it, do it twice for each descendant. Generate $P_N-P_I-P_D-P_A$ such descendants.

Step 7: Stop judgment: go to step 2 unless the stop criteria is met.

\textbf{Remark:} \begin{enumerate}
                   \item Reflection of crossover in our problem. Crossover is not used in our GA procedure for three reasons: (1) inability to remain the number of 1s unchanged in each individual. (2) not able to take the real connotation of our problem, it means nothing to crossover two individuals, especially compared to Step 4. (3) it may incur a scenario that two turbines are too close to each other.
                  % \item Step 4 means that relocate the worst efficient wind turbine to 20 possible positions.
                   \item The inclusion of aliens in Step 5 is to avoid the local optimal.
                   \item Mutation in Step 6 usually means relocate one of the 16 turbines to other possible position.
                  \end{enumerate}

\section{Numerical results}
In this section, we experiment the proposed AGA method for an optimal layout of 16 turbines over 441 possible positions, as discussed in Section \ref{sectref}. We investigate 3 cases: case 1 with a fixed free wind speed of 12 m/s at a single direction; case 2 with a fixed free wind speed of 20 m/s from a single direction; while case 3 deals with variant wind speed with multiple wind directions taken from a typical offshore wind distribution model \cite{Rebecca}.

\subsection{case 1:} 
\begin{figure}
\begin{center}
\includegraphics[width=2.5in]{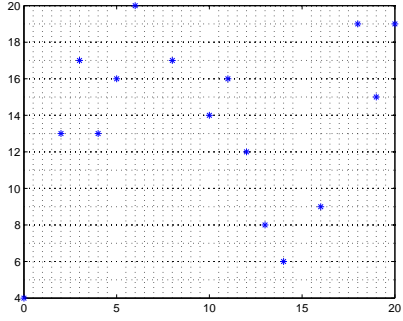}
\setlength\intextsep{1.25\baselineskip plus 1pt minus 1pt}
\caption{Optimal wind farm layout of case 1 with $v=12m/s$  (100\% efficiency achived)}
\label{singleSD}
\end{center}
\end{figure}

\begin{figure}
\begin{center}
\setlength\intextsep{1.25\baselineskip plus 2pt minus 4pt}
\includegraphics[width=2.5in]{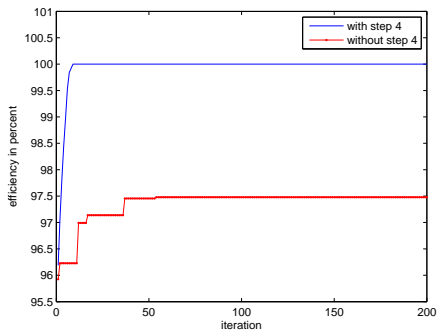}
\caption{Comparison of onvergence performance: adaptive GA vs. conventional GA}
\label{converge}
\end{center}
\end{figure}

The reason why we choose the wind speed of 12 m/s is because it is located in the uprising area of power curve. It has more sensitivity of the wake impact influenced output power. The optimal layout is shown in Fig.\ref{singleSD}. However, if Step 4 is replace with the operation of generating individuals randomly, the converge speed will be much lower and it can almost run forever to get an optimal result as good as with Step 4. Although it is apparent that some turbines are in the same column, it is not easy for the general GA to move the one under wake effect out of it. As is illustrated in Fig.\ref{converge}, our proposed GA algorithm will converge to its 100\% efficiency with only less than 15 steps; while the conventional GA (without Step 4) takes about 60 steps to reach 97.5\% then becomes stuck in a local optimal.
%\begin{figure*}[!t]
%\centering
%\subfloat[Case I]{\includegraphics[width=2.5in]{singleSD4}%
%\hfil
%\subfloat[Case II]{\includegraphics[width=2.5in]{singleSD}%
%\label{fig_second_case}}
%\caption{Simulation results.}
%\label{fig_sim}
%\end{figure*}
%
%\begin{figure}
%\begin{center}
%\includegraphics[width=2.7in]{singleSD4}
%\caption{optimal layout without step 4 (bad performance)}
%\label{singleSD4}
%\end{center}
%\end{figure}
%

\subsection{case 2:} 
\begin{figure}
\begin{center}
\setlength\intextsep{1.25\baselineskip plus 2pt minus 4pt}
\includegraphics[width=2.5in]{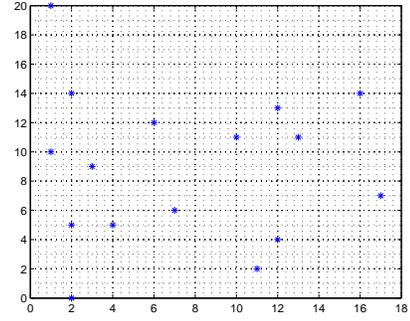}
\caption{Optimal wind farm layout of case 2 with $v= 20$ $ m/s$}
\label{singleSD20}
\end{center}
\end{figure}

When the wind speed is 20 m/s, the optimal solution space is larger than that of $v=12 m/s$. For example, the AGA algorithm locates an optimal layout as illustrated in Fig.\ref{singleSD20}. It is obvious that some turbines are in the wake of others, but the speed of the turbines in the wake is still larger than 14m/s, thus an easily 100\% efficiency can still be achieved even with exisitng wakes.

\subsection{case 3:} 
\begin{figure}
\begin{center}
\setlength\intextsep{1.25\baselineskip plus 2pt minus 4pt}
\includegraphics[width=2.5in]{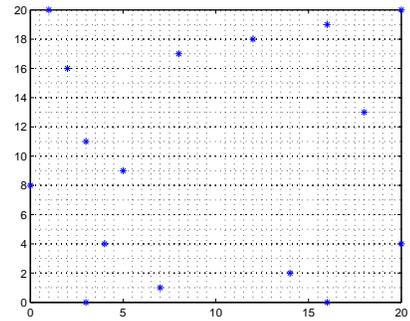}
\caption{Optimal wind farm layout of case 3 with $v= 12$ $ m/s$ and from multiple direction: Efficiency is 97.24\%}
\label{multidirection}
\end{center}
\end{figure}

When the wind speed is 12 m/s and the wind direction is uniformly distributed from $[0^o - 360^o]$. For example, the AGA algorithm locates an optimal layout as illustrated in Fig.\ref{multidirection}. The efficiency is 97.24\%.

\subsection{case 4:} 
\begin{figure}
\begin{center}
\setlength\intextsep{1.25\baselineskip plus 2pt minus 4pt}
\includegraphics[width=2.5in]{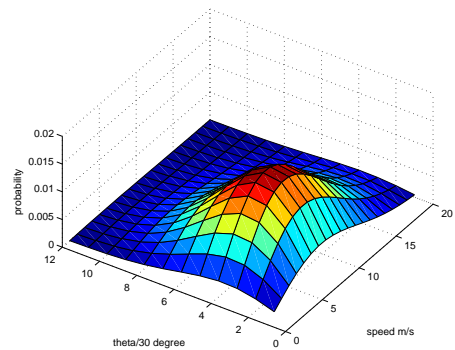}
\caption{Offshore wind speed and direction distribution function $f_w(\theta ,v)$}
\label{distribution}
\end{center}
\end{figure}

\begin{figure}
\begin{center}
\setlength\intextsep{1.25\baselineskip plus 2pt minus 4pt}
\includegraphics[width=2.5in]{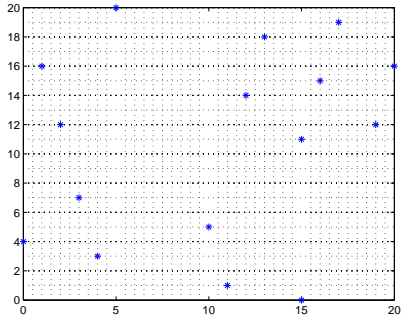}
\caption{Optimal wind farm layout of case 4 with offshore wind distribution }
\label{multiple}
\end{center}
\end{figure}

Variant wind speed and wind direction. According to the real data from Horns Rev and Nysted offshore wind farm, the wind speed distribution fit well to a Weibull distribution where the shape factor is 2.1-2.2 \cite{Rebecca}.We use 2.1 to generate the distribution of the wind speed and direction Fig.\ref{distribution}. The optimal layout in case of multiple wind speed and direction is given in Fig.\ref{multiple}. The achieved efficiency is 97.67\%.  The decrease in achieved efficiency comes from the multiple wind direction and resultng wake effect impact cannot be fully avoided completely.

\section{Looking for The most efficient area: a different perspective}
In this section, we are trying to find an effective way to characterize the most economic area for the wind farm. As you can see that Case 1 and case 2 can achieve 100\% efficiency, so a natural question will be proposed: if the efficiency is 100\%, why not shrink the size of the wind farm and see whether the efficiency is still 100\%, then we will seek how the efficiency drops in the process of shrinking the wind farm. This question is based on a totally different perspective from all the previous research on the optimal layout problem of the wind farm. It is in the hope of giving some insights on how much wind farm areas should be allocated for a given amount of power to be generated. In the next Fig.\ref{avgvsedge}, you can see the decrease of overall power accords with the reduction of the edge size of each cell in the wind farm. The settings of the Fig.\ref{avgvsedge} is multiple direction and multiple wind speed, as the same as the case 4.

\begin{figure}
\begin{center}
\includegraphics[width=2.5in]{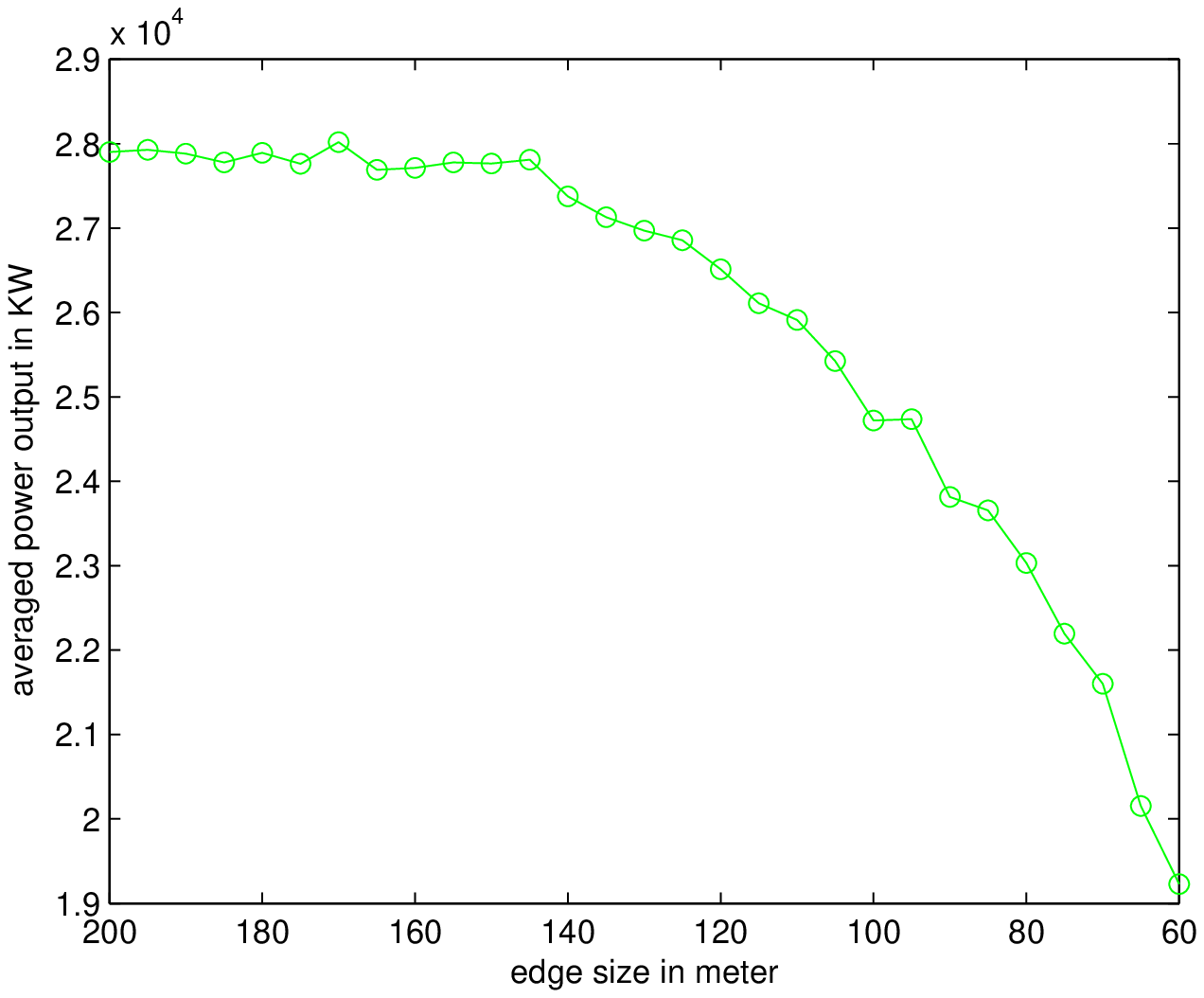}
\caption{averaged power output with different edge size}
\label{avgvsedge}
\end{center}
\end{figure}
\begin{figure}
\begin{center}
\includegraphics[width=2.5in]{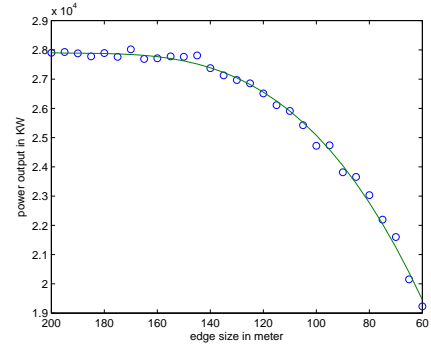}
\caption{3-order polynomial fit with the output power}
\label{polyfit}
\end{center}
\end{figure}

Next we use a three order polynomial to fit the output power curve, as is shown by Fig.\ref{polyfit}

\begin{figure}
\begin{center}
\includegraphics[width=2.5in]{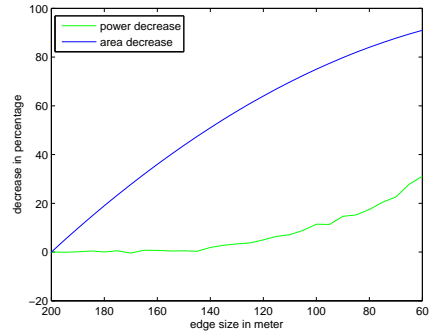}
\caption{Decrease of area vs decrease of power output in percentage}
\label{decrease}
\end{center}
\end{figure}

Then we study the decrease of both power and area in percentage. As is illustrated by Fig.\ref{decrease}. As we can see, when $x=145m$, the total farm area will drop 47.44\% and the power output power will drop only 0.325\%, which is basically unchanged. If we set the allowed power drop to 5\%, the smallest possible area can be decreased to 64\% decrease, which means that we may sacrifice 5\% of power drop compared to the original overall power output in replace of a saving of 64\% of the total area.

\begin{figure}
\begin{center}
\includegraphics[width=2.5in]{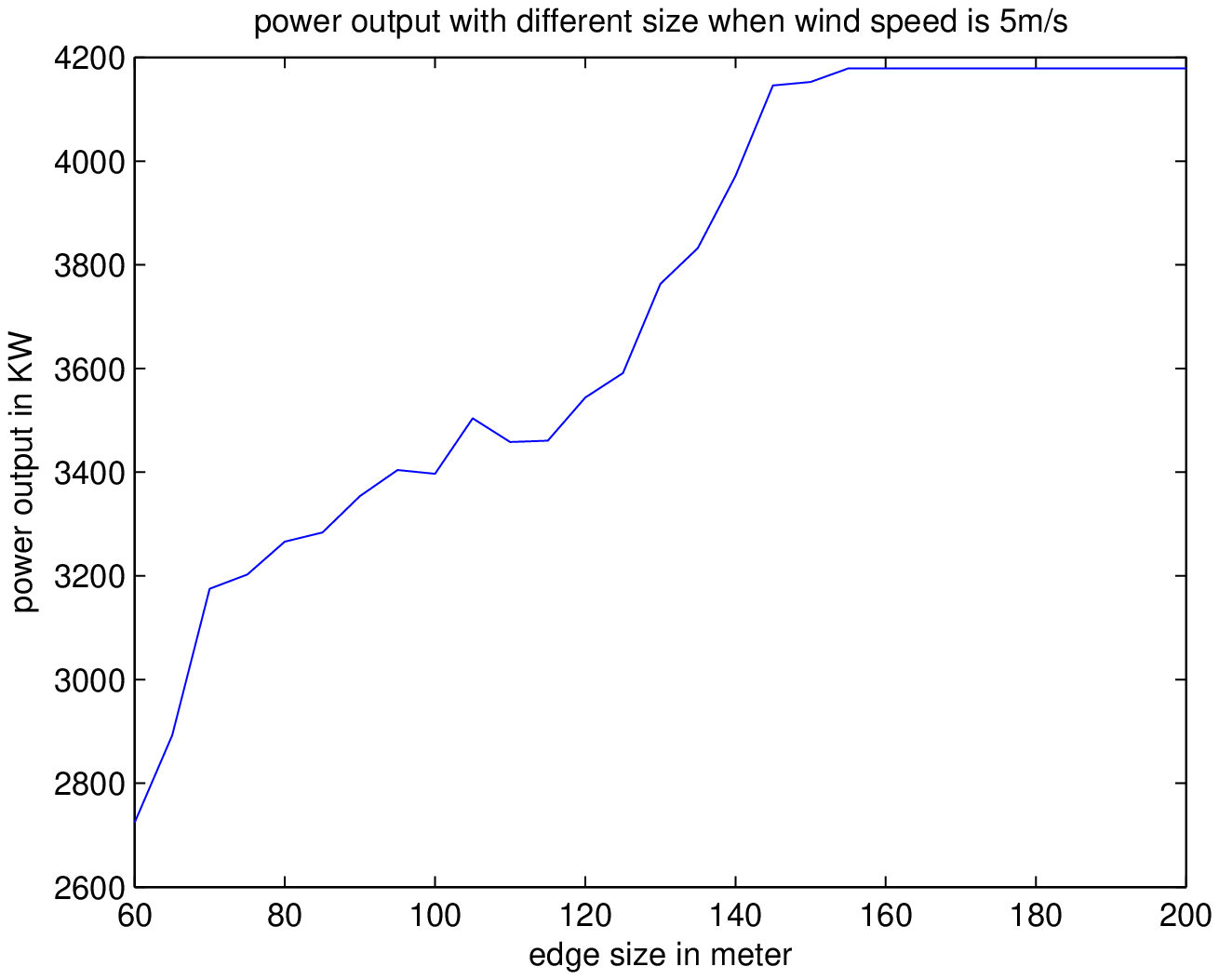}
\caption{averaged power output with different edge size}
\label{singleSDShrink}
\end{center}
\end{figure}

The single wind direction and the single wind speed at speed $5m/s$ is considered. The shrinking of edge also have an impact on the output power as is in Fig.\ref{singleSDShrink}
 In this Fig.\ref{singleSDShrink}, when the edge length is $160m$, the output power remained the same. It means that a reduction of $36\%$ of possible area will not bring about any loss of the total power.
 
\begin{figure}
\begin{center}
\includegraphics[width=2.5in]{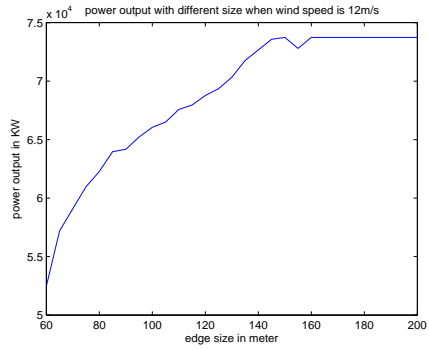}
\caption{power output with different size when wind speed is 12m/s}
\label{shrink12}
\end{center}
\end{figure}

 The single wind direction and the single wind speed at speed $12m/s$ is considered. The shrinking of edge also have an impact on the output power as is in Fig.\ref{shrink12}

\section{The comparison of the layout by AGA to the uniform layout}

\begin{figure}
\begin{center}
\includegraphics[width=2.5in]{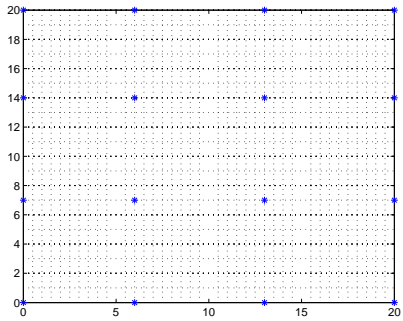}
\caption{uniform layout}
\label{uniformlayout}
\end{center}
\end{figure}

In this paper, we compare the AGA performance with the uniform layout in a wind farm. The uniform layout is given in Fig.\ref{uniformlayout}. All the wind turbines are lined up in a straight line.

\begin{figure}
\begin{center}
\includegraphics[width=2.5in]{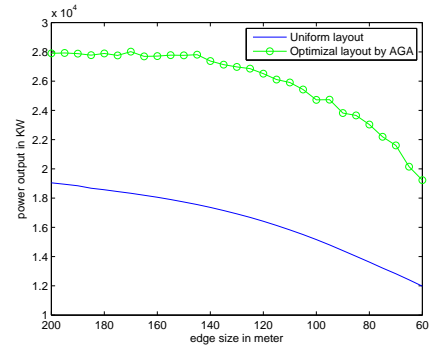}
\caption{averaged power output with different edge size}
\label{uniformvsaga}
\end{center}
\end{figure}

However, the averaged power output in the uniform layout is much lower than those optimized with our proposed AGA. As see in Fig.\ref{uniformvsaga}, our proposed method exceeds in the performance compared to the uniform layout.

\section{Conclusion}
In this paper, we talked in details some considerations when planning the layout of an offshore wind farm, like the fixed number or flexible number of turbines as the optimal parameter, the usefulness of crossover in GA. A new approach is proposed and it is shown to speed up the convergence rate in three different cases. In our novel algorithm, we find the turbine with the worst performance and relocate it, thus guarantee the performance is getting better, at least not getting worse, in each iteration. Next, we studied the impact of wind farm area on the output of the total electricity power. Out research will give some insight on the area determination of a wind farm.

\bibliographystyle{ieeetr}
\bibliography{ref}

\end{document}